# Maximum Entropy and the Glasses You are Looking Through

Peter Grünwald*
EURANDOM
P.O. Box 513
5600 MB Eindhoven
The Netherlands

## Abstract

We give an interpretation of the Maximum Entropy (MaxEnt) Principle in game-theoretic terms. Based on this interpretation, we make a formal distinction between different ways of *applying* Maximum Entropy distributions. MaxEnt has frequently been criticized on the grounds that it leads to highly representation dependent results. Our distinction allows us to avoid this problem in many cases.

## 1 INTRODUCTION

The Maximum Entropy Principle (Jaynes, 1989) is an often successful yet controversial method for inductive inference. It has been justified and criticized in many different ways (Jaynes, 1989; Grove et al., 1994; Halpern & Koller, 1995). Here we give a novel game-theoretic justification that is fundamentally different from previous ones: we show that the MaxEnt distribution for a given constraint is the distribution that minimizes the worst-case expected loss when used for prediction in a certain game. We give several interpretations of this game. We argue that the game-theoretic interpretation is more natural than the usual one, and that it sheds new light on the circumstances in which MaxEnt can be fruitfully applied. Specifically, there are applications of MaxEnt where the same inference problem may be associated with several different games with different worst-case optimal strategies. We use this insight to formally distinguish between qualitatively different ways of *applying* a MaxEnt distribution, ranging from 'completely safe' to 'completely untrustworthy' applications. This also leads to a partial solution of Bertrand's paradox, i.e. the representation dependency of MaxEnt inferences. Sections 2-4 introduce notation and review MaxEnt and the representation dependency problem. Section 5 gives our game-theoretic reinterpretation. Sections 6-8 show how the reinterpretation can be used to distinguish between different ways of applying MaxEnt and to (sometimes) avoid Bertrand's paradox.

*Also: CWI, Kruislaan 413, 1098 SJ Amsterdam, The Netherlands. URL: http://robotics.stanford.edu/~grunwald. Most of the work reported here was done while the author was a postdoctoral fellow at the Robotics Lab, Computer Science Dept., Stanford University, Stanford CA 94305. At the time the author was supported by a TALENT-grant awarded by the Netherlands Organization for Scientific Research (NWO). A very preliminary version of some of the work reported here appeared in (Grünwald, 1998).

## 2 PRELIMINARIES

Consider a finite sample space $\Omega$. We reserve the use of random variable $X$ to denote outcomes in $\Omega$. All other random variables can be vector valued, i.e. they are by definition functions from $\Omega$ to $\mathbf{R}^k$ for some $k > 0$. For random variable $Y : \Omega \to \mathbf{R}^k$, we define the *range* of $Y$, denoted by $\Omega_Y$, as

$$\Omega_Y = \{y \in \mathbf{R}^k \mid \exists x \in \Omega : Y(x) = y\}.$$

By this notation $\Omega_X = \Omega$. We let $\mathcal{P}_Y$ stand for the family of all probability distributions over $\Omega_Y$. For a $P_Y \in \mathcal{P}_Y$ and $\mathcal{A}_Y \subseteq \Omega_Y$, $P_Y(\mathcal{A}_Y)$ denotes the probability mass of $\mathcal{A}_Y$ under $P_Y$. For distributions $P_X \in \mathcal{P}_X$, the notation $P_X(Y = y)$ is short for $P_X(\{x \in \Omega_X \mid Y(x) = y\})$. Let $P_Y \in \mathcal{P}_Y$ and $P_Z \in \mathcal{P}_Z$ be distributions over $\Omega_Y$ and $\Omega_Z$ respectively. We say that $P_Y$ and $P_Z$ are *compatible* (with underlying space $\Omega_X$) if there exists a $P_X \in \mathcal{P}_X$ such that for all $y \in \Omega_Y, P_X(Y = y) = P_Y(\{y\})$ and for all $z \in \Omega_Z, P_X(Z = z) = P_Z(\{z\})$. Intuitively, $P_Y$ and $P_Z$ are compatible if they can be thought of as marginal distributions of a single distribution $P_X$ defined over the more fine-grained space $\Omega_X$. We frequently use random variables that are *indicator functions*. The indicator function for event $\mathcal{A} \subseteq \Omega_X$ is denoted by $\mathbf{1}_{X \in \mathcal{A}}$ and defined by $\mathbf{1}_{X \in \mathcal{A}} = 1$ if $X \in \mathcal{A}$ and 0 otherwise. A *measure* for $\Omega_Y$ is a function $M_Y : \Omega_Y \to (0, \infty)$. It is extended to arbitrary events $\mathcal{A}_Y \subseteq \Omega_Y$ by $M(\mathcal{A}_Y) := \sum_{y \in \mathcal{A}_y} M_Y(y)$. 'Compatibility' of measures $M_Y$ and $M_Z$ is defined analogously to compatibility of probability distributions $P_Y$ and $P_Z$. The *entropy* of a distribution $P$ over $\Omega_X$ relative to measure $M$ over $\Omega_X$ is defined as

$$\mathcal{H}_M(P) := E_P[-\ln \frac{P(X)}{M(X)}] = \sum_{x \in \Omega_X} P(x) \ln \frac{M(x)}{P(x)}.$$

**Notational Convention** If a distribution, (or measure, or set of distributions) is denoted with a subscript $Y$ for some random variable $Y$, we mean a distribution (measure, set of distributions) over $\Omega_Y$ (examples are $P_Y, M_Y, \mathcal{C}_Y$). If a distribution (measure, set of distributions) is denoted *without* subscript, it is always a distribution (measure, set of distributions) over the basic sample space $\Omega_X = \Omega$.

## 3   REVIEW OF MAXENT

Let $\phi_1, \ldots, \phi_k$ be functions from $\Omega_X$ to **R**. In the usual MaxEnt setting, we are given a set of constraints regarding the *expected values* of the functions $\phi_i$ under some unknown distribution $P^*$:

$$E_{P^*}[\phi_1(X)] = t_1, \ldots, E_{P^*}[\phi_k(X)] = t_k \quad (1)$$

where the $t_i$ are values in **R**. By taking the $\phi_i$ to be indicator functions we can express constraints of the form $P^*(Y = y) = t$ for arbitrary random variables $Y$.

We now ask the following question: if the only knowledge we have about $P^*$ are the constraints given by (1), what is then our 'best' guess for $P^*$? (for interpretations of 'best' see Section 6). According to the adherents of maximum entropy we should adopt the distribution $P$ that, among all the distributions satisfying the constraints (1), maximizes the entropy $\mathcal{H}_M(P)$. To formalize this idea we first abbreviate the constraints (1) to

$$E_{P^*}[\phi(X)] = t. \quad (2)$$

Here $\phi(X) = (\phi_1(X), \ldots, \phi_k(X))^T$ is a function from $\Omega_X$ to $\mathbf{R}^k$ and $t$ is a $k$-dimensional vector $(t_1, \ldots, t_k)^T$. Each constraint of form (2) determines a set of probability distributions satisfying the constraint. This set is denoted by $\mathcal{C}$:

$$\mathcal{C} := \{ P \in \mathcal{P}_X \mid E_P[\phi(X)] = t \}. \quad (3)$$

In all our theorems and propositions, we will assume the following

**Regularity conditions** (A) $\Omega_X$ is finite; (B) the set $\mathcal{C}$ mentioned in the theorems is defined as in (3) and is non-empty (this means that every conceivable $\phi$ and $t$ are allowed as long as $\mathcal{C}$ is non-empty).

We can now define maximum entropy inference formally: given a tuple $(\Omega, M, \phi, \mathcal{C})$, where $M$ is a measure over $\Omega$ and $\phi$ and $\mathcal{C}$ are as in (3), MaxEnt tells us to adopt the distribution $P_M^{me}$ given by

$$P_M^{me} := \arg\max_{P \in \mathcal{C}} \mathcal{H}_M(P) = \arg\max_{P \in \mathcal{C}} E_P[-\ln \frac{P(X)}{M(X)}]. \quad (4)$$

Our regularity conditions are sufficient to ensure that a unique $P_M^{me}$ always exists.

## 4   MAXENT, MEASURE AND REPRESENTATION DEPENDENCY

Several versions of MaxEnt and of the related minimum relative entropy principles exist in the literature. By making the entropy $\mathcal{H}_M$ dependent on an underlying measure $M$, we can account for all of these with our Equation 4. We distinguish between two main forms:

**Case A: $M$ NOT available ($U$-MaxEnt)** This is the 'classical' form of MaxEnt for discrete sample spaces (Jaynes, 1989). It does not mention any underlying measure and tells us to pick the distribution $P \in \mathcal{C}$ maximizing $E_P[-\ln P(X)]$. It can be implemented in (4) by taking $M$ to be the *uniform* measure over the sample space $\Omega_X$, defined by $M(x) \equiv 1$. We will refer to this form of MaxEnt as $U$-MaxEnt.

**Case B: $M$ available** We will refer to this case simply as 'MaxEnt'. It has two sub-cases: first, the case where a unique measure ('natural way of counting outcomes') is available a priori. Sometimes, through knowledge of the physics of the domain that is being modeled, one can decide on a unique underlying measure $M$ that is appropriate for the domain at hand (for ways to determine such a measure, see (Jaynes, 1989)). (4) can be directly applied here. Second, the *Minimum Relative Entropy Principle*. This is the case where a prior probability $Q$ over $\Omega_X$ is known, and the goal is to 'update' this prior probability based on the constraint (2). The minimum relative entropy principle tells us to pick the $P$ minimizing $E_P[\ln(P(X)/Q(X))]$. By picking $M = Q$, this can be represented as maximizing entropy relative to $M$.

If a priori knowledge about the domain other than the given constraint (2) is completely lacking, then $U$-MaxEnt (case A) is the only form we can apply. Unfortunately, case A is also the most problematic by far. In contrast to case B, case A typically gives results that are highly *representation dependent*: if the same domain is represented in a different language, MaxEnt may lead to different results. Since the choice of representation seems arbitrary, the results one obtains using $U$-MaxEnt seem arbitrary as well. This fact is – at least in the case of continuous data – often referred to as *Bertrand's Paradox*.

**Example 1 (a simple Bertrand's Paradox)** Let $\Omega_X = \{1, 2, 3\}$ and let there be no further constraints (in our formulation, this can be expressed by picking $\phi(x) \equiv 0$ and constraint $E_{P^*}[\phi(X)] = 0$). Consider an agent (call him Mr. $X$) who wants to infer a distribution over $\Omega_X$ and who uses $U$-MaxEnt. Hence he picks $P_{U_X}^{me}$ as given by (4) with $U_X$ the uniform measure over $\Omega_X$. As is well known, the resulting $P_{U_X}^{me}$ is the uniform distribution over $\Omega_X$. Now consider another agent (Mrs. $Y$) who looks at the same domain at a coarser level. Specifically, she can only distinguish between the case where $X = 1$ on the one hand and $X \in \{2, 3\}$ on the other hand. Mrs. $Y$'s sample space is therefore $\Omega_Y = \{\{1\}, \{2, 3\}\}$. If Mrs. $Y$ uses $U$-MaxEnt she will adopt a uniform measure $U_Y$ over $\Omega_Y$. She will then infer a distribution $P_{U_Y}^{me}$ uniform over $\Omega_Y$. Then $P_{U_Y}^{me}(\{1\}) = 1/2$, $P_{U_X}^{me}(\{1\}) = 1/3$: the distributions inferred by Mr. $X$ and Mrs. $Y$ are incompatible, even though they are based on the same domain. For more subtle examples of this phenomenon, see for example (Halpern & Koller, 1995). In our formulation (with underlying measure), Bertrand's paradox can be equivalently expressed as the dependency of $P_M^{me}$ on the choice of underlying measure $M$. In our example, if Mrs. $Y$ uses the measure $M'_Y$ defined by $M'_Y(\{2, 3\}) := 2$ and $M'_Y(\{1\}) := 1$ then she will infer $P_{M'_Y}^{me}(\{1\}) = 1/3$ after all: this change of measure over $\Omega_Y$ has the same effect on the probability assignments to elements of $\Omega_Y$ as the representation change from $\Omega_Y$ (with measure $U_Y$) to the more fine-grained $\Omega_X$ (with measure $U_X$). This observation will be made precise in Theorem 2, Section 7.

*Important* In many cases, physical background knowledge provides a 'natural' space for representing the domain at hand. For example, if we are to investigate the probabilities of the faces of a (possibly loaded) die, then by symmetry considerations, we should not distinguish a priori between the six faces. It is then only natural to take as basic sample space the space with exactly one outcome for each face, and to take a uniform measure over this space. The representation dependency should be considered problematic only if there is no preferred 'natural' sample space or (equivalently, by Theorem 2), no natural underlying measure/prior.

Some people do not see the above example as problematic: the two agents are facing different 'experimental situations', so it is not so strange that they obtain different results. But then the question is: *what exactly constitutes an 'experimental situation'?* It is this question we will partially answer through our game-theoretic reinterpretation of MaxEnt, which we proceed to discuss.

## 5 MAXENT AS A GAME

The *information inequality* (Cover & Thomas, 1991) tells us that for all distributions $P$ and $Q$ over $\Omega_X$,

$$E_P[-\ln P(X)] \leq E_P[-\ln Q(X)], \quad (5)$$

with equality iff $P = Q$. This implies $\inf_{Q \in \mathcal{P}_X} E_P[-\ln(Q(X)/M(X))] = E_P[-\ln(P(X)/M(X))]$ and hence entropy can be characterized as: $\mathcal{H}_M(P) = \inf_{Q \in \mathcal{P}_X} E_P[-\ln(Q(X)/M(X))]$. The maximum attainable entropy for distributions in a set $\mathcal{C}$ is therefore given by

$$\sup_{P \in \mathcal{C}} \mathcal{H}_M(P) = \sup_{P \in \mathcal{C}} \inf_{Q \in \mathcal{P}_X} E_P[-\ln \frac{Q(X)}{M(X)}]. \quad (6)$$

Readers familiar with game theory (see e.g. (Berger, 1985)) will recognize (6) as the maximin gain of a two-player zero-sum game. If they are acquainted with Von Neumann's minimax theorem, they may further suspect that the following equality holds:

$$\inf_{Q \in \mathcal{P}_X} \sup_{P \in \mathcal{C}} E_P[-\ln \frac{Q(X)}{M(X)}] =$$

$$\sup_{P \in \mathcal{C}} \inf_{Q \in \mathcal{P}_X} E_P[-\ln \frac{Q(X)}{M(X)}] = \mathcal{H}_M(P_M^{me}). \quad (7)$$

This equality indeed holds under very mild conditions, although this does not follow *directly* from Von Neumann or Nash's theorems (which cannot handle arbitrary convex sets of mixed strategies such as $\mathcal{C}$):

**Theorem 1** *Under the regularity conditions of Section 3, Equation 7 holds. Moreover, (a) $\sup_{P \in \mathcal{C}} \inf_{Q \in \mathcal{P}_X} E_P[-\ln(Q(X)/M(X))]$ is reached for (and only for) $P = P_M^{me}$; (b) $\inf_{Q \in \mathcal{P}_X} \sup_{P \in \mathcal{C}} E_P[-\ln(Q(X)/M(X))]$ is reached for (and only for) $Q = P_M^{me}$. Hence (c) we have*

$$P_M^{me} = \arg\min_{P \in \mathcal{P}_X} \sup_{P^* \in \mathcal{C}} E_{P^*}[-\ln \frac{P(X)}{M(X)}], \quad (8)$$

*(d) $P_M^{me}$ is an 'equalizer strategy', i.e. for all $P^* \in \mathcal{C}$,*

$$E_{P^*}[-\ln \frac{P_M^{me}(X)}{M(X)}] = E_{P_M^{me}}[-\ln \frac{P_M^{me}(X)}{M(X)}] \quad (9)$$

A similar theorem with much less conditions on $\Omega_X$ and $\mathcal{C}$ will be provided in (Grünwald & Dawid, 2000).

**Basic Interpretation** Consider the decision-theoretic setting where an Agent has to make decisions about the outcomes in some space $\Omega_X$. Agent's decisions come from a decision space $\mathcal{D}$ and the loss is measured by some function $\text{LOSS} : \Omega_X \times \mathcal{D} \to \mathbf{R} \cup \{\infty\}$. After making a decision $\delta \in \mathcal{D}$, the actual outcome $x \in \Omega_X$ is revealed and Agent incurs a loss $\text{LOSS}(x, \delta)$. Sometimes the decisions $\delta$ are best interpreted as *predictions* of the values of $x$, sometimes they are best interpreted as *game playing strategies*.

The *logarithmic loss function* is a loss function that occurs in several games with several interpretations (Berger, 1985; Cover & Thomas, 1991). The set $\mathcal{D}$ of available decisions for these games consists of all functions $P : \Omega_X \to [0, 1]$ such that $\sum_{x \in \Omega_X} P(x) = 1$. Hence $\mathcal{D}$ is formally equivalent to $\mathcal{P}_X$. However, as we will see, the elements of $\mathcal{D}$ sometimes have interpretations very different from probability distributions. The logarithmic loss function (relative to measure $M$) is defined by $\text{LOSS}(x, P) = -\ln(P(x)/M(x))$ for each $x \in \Omega_X$ and $P \in \mathcal{D}$. Consider now a game where Nature chooses a 'true' distribution $P^*$ and Agent wants to minimize his expected logarithmic loss. If Agent knew $P^*$, he would choose $\arg\min_{P \in \mathcal{P}_X} E_{P^*}[\text{LOSS}(X, P)]$. By the information inequality (5) we see that this is given by $P = P^*$. But now consider the case where Agent only knows that $P^* \in \mathcal{C}$ for some set $\mathcal{C}$. He may now want to minimize his *worst-case (maximal) expected logarithmic loss* over all choices of Nature. This is exactly what is expressed by (8): *the maximum entropy distribution is the worst-case optimal distribution for predicting outcomes of $\Omega_X$ when loss is measured by the logarithmic loss function.*

Why would Agent at all be interested in minimizing logarithmic loss? This game has several important interpretations. Below we discuss one that is of specific interest in the remainder of this paper; others are summarized in Section 5.2.

### 5.1 KELLY GAMBLING INTERPRETATION

Imagine a lottery where there are tickets for sale for betting on outcomes in $\Omega_X = \{1, \ldots, m\}$. Ticket $j$ (for $1 \leq j \leq m$) pays $b$ units if outcome $j$ actually occurs; otherwise, it pays nothing. All tickets cost 1 unit, so all outcomes share the same odds. Agent has some capital $K$ which he wants to invest in lottery tickets. Suppose that Agent thinks that the actual outcomes are distributed according to some distribution $P^*$. Agent's *gambling strategy* can be described by a vector $P = (P(1), \ldots, P(m))$ where $P(j)$ is the fraction of capital $K$ that Agent invests in outcome $j$. That is, he buys $P(j) \cdot K$ tickets for outcome $j$; for convenience we allow buying a non-integer amount of tickets. If Agent plays the game only once, then his expected gain $E_{P^*}[bKP(X)]$ is maximized by the strategy with $P(j) = 1$ for the $j$ with maximum probability $P^*(j)$. But now suppose Agent plays the same game several times. After each round, he reinvests his remaining capital by buying tickets for the next round. So after round 1, his capital is $K_1 = P(x_1)bK$ where $x_1$ is the actual outcome at round 1. After round 2, his capital is $K_2 = P(x_2)bK_1$ etc. If the number of rounds $n$ is not too small or if it is not known in advance how many rounds there will be, it becomes better for Agent to adopt a fundamentally different strategy, sometimes called *proportional gambling* or the *Kelly gambling scheme* (Cover & Thomas, 1991, Chapter 7). This is defined as the gambling strategy $P$ maximizing $E_{P^*}[\ln P(X)]$. This quantity may be interpreted as the *expected growth rate* of the invested capital. To see why this is a sensible strategy, let $P$ and $Q$ be two distributions such that $E_{P^*}[\ln P(X)] > E_{P^*}[\ln Q(X)]$. Suppose the game is played $n$ times, and outcomes $X_1, \ldots, X_n$ are all independently distributed $\sim P^*$. Then, by the strong law of large numbers, $(1/n) \sum_{i=1}^{n} \ln P(X_i) \to E_{P^*}[\ln P(X)]$ and $(1/n) \sum_{i=1}^{n} \ln Q(X_i) \to E_{P^*}[\ln Q(X)]$ with probability 1. It follows that there exists an $\epsilon > 0$ such that with probability 1, for all large $n$, $\sum_{i=1}^{n} \ln P(X_i) > \sum_{i=1}^{n} \ln Q(X_i) + n\epsilon$. This implies

$$\prod_{i=1}^{n} P(X_i) > e^{n\epsilon} \prod_{i=1}^{n} Q(X_i), \qquad (10)$$

with $P^*$-probability 1, for all $n$ larger than some $n_0$. If Agent uses strategy $P$ at each round $i$, his capital after $n$ rounds is given by $Kb^n \prod_{i=1}^{n} P(x_i)$. Together with (10) this implies that for any two strategies $P$ and $Q$, with $P^*$ probability 1, Agent's end capital is exponentially larger for the strategy with larger expected growth rate. So (at least if $n$ is large or unknown) a rational Agent should adopt the strategy $P$ maximizing $E_{P^*}[\ln P(X)]$. If the only thing Agent knows about $P^*$ is that $P^* \in \mathcal{C}$, it is a good idea for the Agent to maximize his *worst-case expected growth rate*, i.e. to pick the strategy $P$ that maximizes $\min_{P^* \in \mathcal{C}} E_{P^*}[\ln P(X)]$, which is identical to the distribution minimizing $\max_{P^* \in \mathcal{C}} E_{P^*}[-\ln P(X)]$. The latter distribution is the MaxEnt distribution with uniform measure $M$: $P_M^{me}$ *is the worst-case optimal distribution in the Kelly gambling game, maximizing, with probability 1, the worst-case end-capital for all large $n$.* A uniform underlying measure $M$ corresponds to the game with equal odds for all outcomes; non-uniform measures correspond to games with non-uniform odds.

### 5.2 MDL & OTHER INTERPRETATIONS

The minimax game (8) has several other interpretations. We mention two. First, there is a *statistical in-*

*terpretation*: many statistical inference procedures can be interpreted as trying to infer, for a given set of data, the probability model within some class of models $\mathcal{M}$ that is 'as close as possible' to the unknown, 'true' data generating distribution $P^*$. In Maximum Likelihood and some Bayesian inference procedures, 'closeness' is measured by means of the *Kullback-Leibler* KL-distance, which, for fixed $P^*$ and $M$, only differs from the expected logarithmic loss $E_{P^*}[-\ln(P(X)/M(X))]$ by a constant. Second, there is an interpretation in terms of the *Minimum Description Length (MDL) Principle* (Grünwald, 1998), a method for inductive inference that is based on data compression. MDL can be seen as a mathematical formalization of Occam's Razor. It turns out that based on the minimax formulation (8), MaxEnt can be interpreted as a form of MDL; this is shown in (Grünwald, 1998). Finally, we note that even if we do not assume the existence of a unique true distribution $P^*$ (to which for example the Bayesians may object) a form of our analysis can still be performed. All this will be treated in detail in the journal version of this paper.

## 6 THE GLASSES YOU ARE LOOKING THROUGH

Let us now stand back and ask why Agent would like to infer a probability distribution over a domain in the first place. Usually, this is because he would like to make good or at least reasonable predictions or decisions concerning some random variable $Y$ referring to the domain (more general prediction tasks will be considered in later sections). If $Y$ is a random variable in the domain and $\mathcal{D}$ is a set of available decisions, then the quality of such decisions is usually measured by some loss function LOSS: $\Omega_Y \times \mathcal{D} \to \mathbf{R} \cup \{\infty\}$. If Agent knew the 'true' distribution $P^*$ governing domain $\Omega_X$, then he could use this knowledge to make optimal predictions for *any* given loss function by predicting using the action $\hat{\delta} = \arg\min_{\delta \in \mathcal{D}} E_{P^*}[\text{LOSS}(Y, \delta)]$ (Berger, 1985).

But the interpretation of $P_M^{me}$ from the minimax point of view (8) suggests that $P_M^{me}$ should first and foremost be interpreted as the strategy to adopt in the game described in Section 5 and *not* as the distribution $P^*$ according to which data are distributed. This leads to a key insight: perhaps we should not regard $P_M^{me}$ as a guess of the 'true' $P^*$ to be used by Agent in every prediction task that can be defined over the domain. It may be better to think of $P_M^{me}$ as being *wrong yet useful* in that it may be a reasonable guess of $P^*$ for use in some possible prediction tasks (i.e. for some random variables and loss functions) but a quite unreasonable guess for use in other combinations of random variables and loss functions[1]. To make this idea concrete, suppose Agent has no access to $P^*$ but approximates it using $P_M^{me}$. For given loss function LOSS, Agent can use $P_M^{me}$ to arrive at a prediction by picking

$$\hat{\delta} = \underset{\delta \in \mathcal{D}}{\arg\min}\ E_{P_M^{me}}[\text{LOSS}(Y, \delta)]. \quad (11)$$

In general, this will lead to reasonable results if $\forall \delta \in \mathcal{D} : E_{P_M^{me}}[\text{LOSS}(Y, \delta)] \approx E_{P^*}[\text{LOSS}(Y, \delta)]$. For each $\delta \in \mathcal{D}$, the function $\psi$ defined by $\psi(x) := \text{LOSS}(Y(x), \delta)$ is a random variable. Hence, in order to be able to decide for arbitrary loss function LOSS whether $P_M^{me}$ as used in (11) will lead to reasonable predictions, it suffices if for arbitrary random variables $\psi : \Omega_X \to \mathbf{R}^k$, we can decide whether $E_{P_M^{me}}[\psi(X)]$ is a reasonable guess for $E_{P^*}[\psi(X)]$. Let us analyze this question further.

In order to predict the likely value of $E_{P^*}[\psi(X)]$, Agent must assign probabilities to the values in $\Omega_\psi$ that random variable $\psi$ takes. This involves 'looking at the domain' in terms of the function $\psi$; in other words, $\psi$ determines the 'glasses' through which Agent looks at data $X$ from sample space $\Omega_X$. But when *inferring* $P_M^{me}$, Agent observes averages of values of the function $\phi$. Hence Agent looks at the world in terms of $\Omega_\phi$. As discussed in Section 5.1, $P_M^{me}$ maximizes the worst-case gain when used for Kelly Gambling *on the outcomes in $\Omega_X$ against odds determined by $M$*. We may now ask ourselves why Agent should be interested in a distribution that maximizes gain when betting on outcomes in $\Omega_X$ if both the observables $(\phi(X))$ and the 'predictables' $(\psi(X))$ are outcomes in spaces different from $\Omega_X$. If no a priori measure $M$ is available, $U$-MaxEnt advocates a uniform $M$. But should Agent adopt a uniform $M$ over $\Omega_X, \Omega_\phi$ or $\Omega_\psi$? Indeed $\phi, \psi$ and $X$ may be related in such a way that the optimal gambling strategies against uniform odds for outcomes in $\Omega_X, \Omega_\phi$ and $\Omega_\psi$ are mutually incompatible. This immediately suggests that postulating a uniform measure over $\Omega_X$ may not be the right thing to do; and that it is not so strange that it leads to representation dependence. Analyzing this fact with the game-theoretic interpretation in mind, we find (in Section 8.3) that if $\phi$ and $\psi$ are related in a certain way, we can do something about this. Namely, *as long as $\Omega_\psi$ is at least as 'coarse' as $\Omega_\phi$* ('looking at an outcome $x$ through the 'glasses' $\phi$ allows a view on $\Omega_X$ that is at least as fine-grained as the view through the glasses $\psi$'), it is still possible to postulate an a priori measure $M$ (not necessarily uniform over $\Omega_X$) such that the Kelly gambling games on outcomes in $\Omega_\phi$ and $\Omega_\psi$ and $\Omega_X$ against odds determined by $M$ share the same worst-case optimal strategy $P_M^{me}$. Moreover, as we shall see, postulating $M$ in this way makes $P_M^{me}$ representation independent *as long as it is only used for predictions concerning random variables $\psi$ that are at least as 'coarse' as $\phi$*. Similarly, under stronger conditions on the relation between the functions $\phi$ and $\psi$, one can guarantee not only that $P_M^{me}$ is representation independent but even that $E_{P_M^{me}}[\psi] = E_{P^*}[\psi]$, i.e. that the MaxEnt guess of $E_{P^*}[\psi]$ is correct.

---

[1] Related ideas are quite common in statistics and Machine Learning. As an example, 'Naive Bayes' models are joint probability distributions defined over discrete random variables $X_1, \ldots, X_k, Y$ of a certain parametric form. They usually perform exceedingly well *when used to predict values of $Y$ conditional on $X_1, \ldots X_k$* (Friedman et al., 1997) *under the 0/1 (classification) loss function*. Yet they make all kinds of unwarranted independence assumptions that might lead to disastrous results if they were used to predict, say, the value of $X_2$ conditional on the value of $X_1$.

The moral of the story is that, depending on how the 'glasses' (ways of looking at the data) $\phi$ and $\psi$ are related, applying MaxEnt may (a) be inherently representation dependent (and should therefore not be used at all), or (b) be representation independent but not guaranteed to be 'optimal' or 'correct' (in this case it can be used as an inductive guess) or (c) be guaranteed to lead to correct or optimal predictions (in which case it should certainly be used). In Section 8 we formalize this distinction. We first need to establish the relation between representations and underlying measures.

## 7 REPRESENTATION & MEASURE

In this section we will show that if we stick to a fixed measure, the results given by MaxEnt are invariant under all reasonable representation changes.

For two random variables $Y$ and $Z$ we say that $Y$ *determines $Z$* if there exists a function $g : \Omega_Y \to \Omega_Z$ such that for each $x \in \Omega$, $g(Y(x)) = Z(x)$. In the language of measure theory, this can be simply expressed as '$Z$ is *measurable in the $\sigma$-algebra generated by $Y$*'. More generally, we say that '$Y$ determines $Z$ over $\mathcal{A} \subseteq \Omega_X$' if there exists $g : \Omega_Y \to \Omega_Z$ such that for each $x \in \mathcal{A}$, $g(Y(x)) = Z(x)$. A set $\Omega_V$ is an *underlying space* for space $\Omega_Z$ if there exists a function $h : \Omega_V \to \Omega_Z$ such that for all $z \in \Omega_Z$, there exists $v \in \Omega_V$ with $h(v) = z$. We can think of an underlying space $\Omega_V$ as a space that is at least as or more fine-grained than $\Omega_Z$. If $V$ is a random variable $\Omega_X \to \Omega_V$, then this simply means that $V$ determines $Z$. The notion is more general in that we can take $Z = X$. In this case, $\Omega_V$ provides a new sample space such that all random variables that can be expressed as functions of $\Omega_X$ can also be expressed as functions of $\Omega_V$. Let $\Omega_V$ be an underlying space for $\Omega_X$. For any random variable $Y : \Omega_X \to \mathbf{R}$, we write $Y_V$ to denote the random variable $\Omega_V \to \mathbf{R}$ that corresponds to $Y$ in the underlying space: for all $v \in \Omega_V$, $Y_V(v) := Y(h(v))$ with $h : \Omega_V \to \Omega_X$ defined as above. With this convention, the function $h$ itself can be written as $h \equiv Y_V$. More generally, let $W_V$ and $T_V$ be any two random variables $\Omega_V \to \mathbf{R}$ in the underlying space $\Omega_V$, such that $W_V$ determines $T_V$. Then $T_W$ is defined as a function $\Omega_W \to \mathbf{R}$ with $\forall v \in \Omega_V : T_W(W_V(v)) := T_V(v)$ (in other words, $T_W$ is a random variable in the intermediate space $\Omega_W$ that always takes on the same value as $T_V$). With this notation, $\phi$ can be equivalently written as $\phi_X$.

A MaxEnt problem with given underlying measure is characterized by a tuple $(\Omega_X, M_X, \phi_X, \mathcal{C}_X)$ (Section 3). A *valid representation shift* of MaxEnt problem $(\Omega_X, M_X, \phi_X, \mathcal{C}_X)$ is a triple $(\Omega_V, \Omega_W, M_W)$. Here $\Omega_V, \Omega_W$ are sets and $M_W$ is a measure over $\Omega_W$ satisfying

1. $\Omega_V$ is an underlying space both for $\Omega_X$ and for $\Omega_W$;
2. the random variable $W_V$ determines $\phi_V$;
3. $M_W$ and $M_X$ are compatible (with underlying space $\Omega_V$; see Section 2).

$\Omega_X$ is called the *original representation space* and $\Omega_W$ is called the *new representation space*.

Intuitively, a representation shift is 'valid' if the constraint $E_{P^*}[\phi] = t$ can still be expressed in the new space. A valid representation shift $(\Omega_V, \Omega_W, M_W)$ induces a new MaxEnt problem $(\Omega_W, M_W, \phi_W, \mathcal{C}_W)$ defined as follows:

$$P^{me}_{W;M_W} := \arg\max_{P_W \in \mathcal{C}_W} E_{P_W}[-\ln \frac{P_W(W_W)}{M_W(W_W)}],$$

where $\mathcal{C}_W$ is the set of all distributions $P_W$ over $\mathcal{P}_W$ that are compatible (with underlying space $\Omega_V$) with some $P_X \in \mathcal{C}$. By Theorem 1 we have that

$$P^{me}_{W;M_W} = \arg\min_{P_W \in \mathcal{P}_W} \sup_{P^*_W \in \mathcal{C}_W} E_{P^*_W}[-\ln \frac{P_W(W_W)}{M_W(W_W)}].$$

The following result is a simple extension of a well-known theorem (see, for example (Shore & Johnson, 1980)). It shows that, if a measure $M_X$ for the original representation space is available, then a valid representation shift leads to the same MaxEnt inferences for all random variables $Y$ that can be expressed both in terms of the original and in terms of the new representation space. In other words, as long as $M_X$ is available, MaxEnt is representation independent.

**Theorem 2** *Let $(\Omega_V, \Omega_W, M_W)$ be a valid representation shift for MaxEnt problem $(\Omega_X, M_X, \phi_X, \mathcal{C}_X)$. Then for all $Y : \Omega_X \to \mathbf{R}$ such that $X$ determines $Y$ and $W_V$ determines $Y_V$, we have for all $y \in \Omega_Y$:*

$$P^{me}_{W;M_W}(Y_W = y) = P^{me}_{X;M_X}(Y_X = y)$$

The present game-theoretic view provides a novel and simple interpretation of this result. Roughly speaking, a representation change amounts to a change in the set of tickets one can buy in the Kelly gambling game (outcomes in $\Omega_W$ rather than $\Omega_X$); by adopting a measure (odds) over $\Omega_W$ that is compatible with the measure (odds) $M_X$ over $\Omega_X$, one ensures that the prices of the different tickets will change along with the representation change so that the gambling game in the new space is essentially equivalent to the original game.

This implies that representation dependence of $U$-MaxEnt stems *only* from the fact that, if we change representation from $\Omega_X$ to $\Omega_W$, we adopt mutually incompatible measures (namely, the uniform measures in both spaces). In fact, we can view each measure $M_X$ (as long as it is rational-valued) as an implicit re-representation of the problem to a different underlying space $\Omega_V$ in which $M_X$ (or, more precisely, the measure over $\Omega_V$ that is compatible with $M_X$) is uniform over $\Omega_V$. We will use of this insight in the next section, where we show how to apply MaxEnt in a careful manner.

## 8 HOW TO APPLY MAXENT

In this section we return to our previous notation, i.e. $P_M^{me}$ is short for $P_{X;M}^{me}$.

Whenever in what follows we say 'we apply MaxEnt for guessing $E_{P^*}[Y|Z]$ relative to measure $M$' this means that we use $P_M^{me}$, defined for a set of constraints $\mathcal{C}$ as given by (3) as follows: we observe the value $z$ taken on by $Z$. Then we infer (guess) that

$$E_{P_M^{me}}[Y|Z=z] \approx E_{P^*}[Y|Z=z]. \quad (12)$$

where $P^*$ is the unknown 'true' distribution. We want to arrive at the general conditions on $\phi, Y$ and $Z$ under which inference (12) can be expected to be a reasonable guess and we want to know what Agent should do if he does not know what $M$ to pick. To treat this question in full generality, we will assume that Agent uses a *set* $\mathcal{M}$ of 'a priori possible underlying measures'. The case where an underlying measure or prior $M$ is available can now be formulated by setting $\mathcal{M} := \{M\}$. In the case where we have a definite reason to pick $\Omega_X$ as our basic representation space (e.g. the case of throwing dice, see Example 1), we can take $\mathcal{M} := \{U_X\}$, where $U_X$ is the uniform measure over $\Omega_X$. If no underlying measure can be determined at all, we will set $\mathcal{M}$ to be the class of *all* measures over $\Omega_X$. As we will see, this will make MaxEnt undefined in most cases. In order to make it well-defined, we must first guess a subset $\mathcal{M}'$ of $\mathcal{M}$; that case will be treated in Section 8.3.

We are now ready to present our *hierarchy* of different forms of applying MaxEnt.

**Definition 1** *Application of MaxEnt for guessing $E_{P^*}[Y|Z]$ relative to a set of measures $\mathcal{M}$ is called*

1. conditionally correct *if* $E_{P_M^{me}}[Y|Z=z] = E_{P^*}[Y|Z=z]$ *for all* $M \in \mathcal{M}$, *all* $z \in \Omega_Z$, *all* $P^* \in \mathcal{C}$.

2. conditionally calibrated *if there exists a random variable $V$ such that $Z$ determines $V$, and $E_{P_M^{me}}[Y|Z=z] = E_{P_M^{me}}[Y|V_Z=V_Z(z)] = E_{P^*}[Y|V_Z=V_Z(z)]$ for all $M \in \mathcal{M}$, all $z \in \Omega_Z$, all $P^* \in \mathcal{C}$.*

3. well-defined *if* $E_{P_{M_1}^{me}}[Y|Z=z] = E_{P_{M_2}^{me}}[Y|Z=z]$ *for all* $M_1, M_2 \in \mathcal{M}$, *all* $z \in \Omega_Z$.

4. ill-defined *otherwise.*

We note that 1. $\Rightarrow$ 2. $\Rightarrow$ 3. Category 2., while perhaps the most interesting, takes too long to discuss here. It will be explained in detail in the journal version of this paper. Category 4. concerns applications of Maximum Entropy that lead to arbitrary results and hence should be avoided (but see the next section!). By the discussion of the previous section, every (rational-valued) measure $M \in \mathcal{M}$ corresponds to an 'allowed' representation shift. Therefore, ill-defined applications of MaxEnt with respect to the set of all measures over $\Omega_X$ correspond to those applications of $U$-MaxEnt that are representation dependent. The other extreme is Category 1, which concerns applications of Maximum Entropy that are completely without risk. If MaxEnt is 'conditionally correct' for $E_{P^*}[Y|Z=z]$, the guess $E_{P_M^{me}}[Y|Z=z] = E_{P^*}[Y|Z=z]$ is guaranteed to be correct and involves no inductive inference. Theorem 3 below determines when this is the case.

### 8.1 CONDITIONAL CORRECTNESS

Let $\psi : \Omega_X \to \mathbf{R}$ and $\phi : \Omega_X \to \mathbf{R}^k$ be two functions and let $\mathcal{A} \subseteq \Omega_X$. We say that $\psi$ *is an affine function of $\phi$ over subdomain $\mathcal{A}$* if there exists $(\alpha_0, \ldots, \alpha_k) \in \mathbf{R}^{k+1}$ such that for all $x \in \mathcal{A}, \psi(x) = \alpha_0 + \sum_{i=1}^k \alpha_i \phi_i(x)$. We define the *support* SUPP of $\mathcal{P}'_X \subseteq \mathcal{P}_X$ by

$$\text{SUPP}(\mathcal{P}'_X) := \{x \in \Omega_X \mid P(x) > 0 \text{ for some } P \in \mathcal{P}'_X\}.$$

**Theorem 3** *If $\psi$ is an affine function of $\phi$ over subdomain* SUPP$(\mathcal{C})$, *then for all $P_1^*, P_2^* \in \mathcal{C}$, $E_{P_1^*}[\psi(X)] = E_{P_2^*}[\psi(X)]$. If $\psi$ is not an affine function of $\phi$ over subdomain* SUPP$(\mathcal{C})$ *then there exist $P_1^*, P_2^* \in \mathcal{C}$ such that $E_{P_1^*}[\psi(X)] \neq E_{P_2^*}[\psi(X)]$.*

Let $Z$ be a 'trivial' random variable, i.e. $\forall x \in \Omega_X, Z(x) = 1$. Theorem 1 implies that for any $M \in \mathcal{M}$ and any $\mathcal{C}$ of form (3), $P_M^{me} \in \mathcal{C}$. Theorem 3 now gives that for arbitrary sets of measures $\mathcal{M}$, applying MaxEnt to guess $E_{P^*}[Y|Z]$ is conditionally correct iff $Y$ is an affine function of $\phi$ over subdomain $\text{SUPP}(\mathcal{C})$. To determine whether applying MaxEnt to guess $E_{P^*}[Y|Z]$ is conditionally correct for non-trivial $Z$ (the case that $Z(x)$ varies over $\text{SUPP}(\mathcal{C})$), we interpret the conditioning event $Z = z$ as the additional constraint $E_{P^*}[\mathbf{1}_{Z=z}] = 1$. Clearly, this constraint is of the required form (2). Let, for $z \in \Omega_Z$,

$$\mathcal{C}^{(z)} = \mathcal{C} \cap \{P^* \in \mathcal{P}_X \mid E_{P^*}[\mathbf{1}_{Z=z}] = 1\}. \quad (13)$$

Then, by the same reasoning as above, applying MaxEnt to guess $E_{P^*}[Y|Z]$ is conditionally correct iff $Y$ is an affine function of $\phi$ over subdomain $\text{SUPP}(\mathcal{C}^{(z)})$ for all $z \in \Omega_Z$. This seems to imply that 'correct' applications of MaxEnt are trivial: if we know that $Y$ is affine in $\phi$, i.e. $Y(x) = \alpha_0 + \sum_{i=1}^{k} \alpha_i \phi_i(x)$, we can also directly infer that $E_{P^*}[Y] = \alpha_0 + \sum_{i=1}^{k} \alpha_i E_{P^*}[\phi_i(X)] = \alpha_0 + \sum_{i=1}^{k} \alpha_i E_{P_M^{me}}[\phi_i(X)] = E_{P_M^{me}}[Y(X)]$ (the second equality follows because both $P^*$ and $P_M^{me}$ are members of $\mathcal{C}$). However, there is at least one case where 'correct' applications are not entirely trivial. From Theorem 1 we see that for all $P^* \in \mathcal{C}$, $E_{P_M^{me}}[-\ln(P_M^{me}(X)/M(X))] = E_{P^*}[-\ln(P_M^{me}(X)/M(X))]$ (indeed, it turns out that $-\ln(P_M^{me}(\cdot)/M(\cdot))$ is an affine function of $\phi$). This means that MaxEnt is 'correct' for inferring the expected logarithmic loss in the games described in Section 5, e.g. in Kelly Gambling with odds determined by $M$. This can be interpreted as follows: imagine an Agent who adopts $P_M^{me}$ and uses it for prediction of outcomes in $\Omega_X$ with respect to log-loss relative to measure $M$. For gambling strategy $\delta \in \mathcal{P}_X$, he expects to incur loss $E_{P_M^{me}}[-\ln \frac{\delta(X)}{M(X)}]$. Agent therefore decides to use the $\delta$ which minimizes this, which by the information inequality (Section 5) is given by $\delta = P_M^{me}$. Using this choice of $\delta$, Agent *expects* to make an average loss of $E_{P_M^{me}}[-\ln(P_M^{me}(X)/M(X))]$. In reality, he will make an average loss of $E_{P^*}[-\ln(P_M^{me}(X)/M(X))]$. Since these two are equal, Agent will have the right idea of how well he will be able to predict on average *even if $P_M^{me}$ is wrong*.

### 8.2 REPRESENTATION DEPENDENCE

By Definition 1, if $\mathcal{M}$ contains only a single element, MaxEnt is well-defined for guessing $E_{P^*}[Y|Z]$ for all $Y$ and $Z$. By Theorem 2, the guesses are also representation independent. The other extreme is the case where no underlying measure is known at all and $\mathcal{M}$ contains all measures over $\Omega_X$. It is straightforward to show

**Proposition 1** *Suppose $\mathcal{M}$ contains all measures over $\Omega_X$. Then MaxEnt is ill-defined for guessing $E_{P^*}[Y|Z]$ iff it is not conditionally correct for guessing $E_{P^*}[Y|Z]$.*

Since different measures correspond to different representations, this corresponds to the fact that $U$-MaxEnt is highly representation dependent. The only way we can get to the interesting cases 2. and 3. of Definition 1 is by restricting the set of available measures $\mathcal{M}$. It seems we have not gained anything so far, since we do not know how this should be done (we have already seen that choosing $\mathcal{M} = \{U_X\}$, with $U_X$ the uniform measure over $\Omega_X$ leads to representation dependence)! But it turns out that if the functions $\phi$, $Y$ and $Z$ are related in certain ways, then we can choose a subset of $\mathcal{M}$ in a different way that does preserve representation independence.

### 8.3 REPRESENTATION INDEPENDENCE

We want to first select a subset $\mathcal{M}'$ of $\mathcal{M}$ and then apply MaxEnt to guess $E_{P^*}[Y|Z]$ in such a way that the whole (2-step) procedure becomes representation independent. Since we have no fixed measure or prior over $\Omega_X$, the choice of $\Omega_X$ as our basic sample space is essentially arbitrary. Therefore our procedure for selecting measures should give the same result for every alternative choice of sample space in which both the constraint $E_{P^*}[\phi] = t$ and the guess $E_{P^*}[Y|Z]$ can be expressed. Our previous analysis suggests a novel way of guessing $\mathcal{M}'$ which in many cases achieves this. It is based on the idea that the 'observables' in our problem are really the outcomes in $\Omega_\phi$ and not the outcomes in $\Omega_X$. This suggests postulating a uniform measure $U_\phi$ over $\Omega_\phi$ *rather than* $\Omega_X$. If we then restrict the functions $\psi$ about which we make guesses to those that are *determined* by $\phi$, the arbitrary choice of our basic representation space $\Omega_X$ becomes irrelevant and we are guaranteed to make the same predictions independent of whatever $\Omega_X$ we choose. We now formalize this idea.

When no underlying measure is given, a MaxEnt problem is determined by a triple $(\Omega_X, \phi_X, \mathcal{C}_X)$. Note that, compared to the treatment in Section 7, the measure $M_X$ is missing. A valid representation shift of such a MaxEnt problem is a pair $(\Omega_V, \Omega_W)$ such that (1) $\Omega_V$ is an underlying space both for $\Omega_X$ and for $\Omega_W$; and (2), the random variable $W_V$ determines $\phi_V$ (again, compare to the definition of valid representation shift in Section 7). The representation shift leads to a new MaxEnt problem $(\Omega_W, \phi_W, \mathcal{C}_W)$ where $\mathcal{C}_W$ is the set of distributions over $\Omega_W$ compatible to $\mathcal{C}_X$ (compatibility with respect to underlying space $\Omega_V$).

Let $(\Omega_V, \Omega_W)$ be any valid representation shift of the MaxEnt problem $(\Omega_X, \phi_X, \mathcal{C}_X)$. Let $U_\phi$ be the uniform measure over $\Omega_\phi$. Let $\mathcal{M}'_X$ be the class of all measures over $\Omega_X$ that are compatible with $U_\phi$, and let $\mathcal{M}'_W$ be the class of all measures over $\Omega_W$ that are compatible with $U_\phi$ (compatibility with respect to underlying space $\Omega_V$).

**Proposition 2 (Bertrand's Paradox 'resolved')**
Let $\psi : \Omega_X \to \mathbf{R}$ be a random variable in the original representation space. If $\phi$ determines $\psi$ on SUPP($\mathcal{C}$), then for all $M_W \in \mathcal{M}'_W$ and all $M_X \in \mathcal{M}'_X$, $E_{P^{me}_{W;M_W}}[\psi_W] = E_{P^{me}_{X;M_X}}[\psi_X]$; if $\phi$ does not determine $\psi$ on SUPP($\mathcal{C}$), there exist $M_X, M'_X \in \mathcal{M}'_X$ such that $E_{P^{me}_{X;M_X}}[\psi_X] \neq E_{P^{me}_{X;M'_X}}[\psi_X]$.

The proposition says that our two-step procedure gives representation independence iff MaxEnt is used to guess functions $\psi$ that are determined by $\phi$. Defining $\mathcal{C}^{(z)}$ as in (13) and reasoning exactly as below Theorem 3, we obtain the following (informally stated)

**Corollary 1** *The procedure of first setting $\mathcal{M}_X$ to be the set of measures over $\Omega_X$ that are compatible with $U_\phi$ and then applying MaxEnt to guess $E_{P^*}[Y|Z]$ relative to set of measures $\mathcal{M}_X$ is a representation independent procedure iff, for all $z \in \Omega_Z$, $\phi$ determines $Y$ over subdomain* SUPP($\mathcal{C}^{(z)}$). *In particular this will be the case if $\phi$ determines both $Y$ and $Z$.*

Summarizing, if all measures $M$ over $\Omega_X$ are a priori possible, then instead of using $U$-MaxEnt it may be better to adopt a uniform measure $U_\phi$ over the space of *observables* $\Omega_\phi$. In this way one obtains a procedure which is representation independent for guessing a large class of random variables $Y$ conditional on a large class of random variables $Z$. The measure $U_\phi$ induces a set $\mathcal{M}_X$ of compatible measures over $\Omega_X$. If $Y$ and $Z$ are such that even for this restricted set of measures, applying MaxEnt to guess $E_{P^*}[Y|Z]$ is ill-defined, then, in our view, any specific guess of $E_{P^*}[Y|Z]$ obtained by a further restriction of $\mathcal{M}'_X$ is basically arbitrary. We feel that in such cases, one should refrain from using MaxEnt altogether.

## 9 FINAL REMARKS

**Non-convex constraints** A major goal for future work is to analyze the behavior of $P^{me}_X$ in minimax terms for constraints that go beyond form (2). For inequality constraints $(E[\phi(X)] \geq t)$, adjusted versions of all our results still hold. For constraints such that $\mathcal{C}$ becomes *non-convex*, MaxEnt is known to lead to rather strange results. Interestingly, for such constraints, Theorem 1 does not apply and the minimax $P^{me}_M$ is not equal any more to the traditional maximin $P^{me}_M$. We suspect that the minimax version gives preferable results in such cases. Consider for example *disjunctive constraints* (Grove et al., 1994): let $\Omega_X = \{0, 1\}$, $M$ uniform, and let the constraint be $[P^*(X = 1) = 0.1] \vee [P^*(X = 1) = 0.95]$. Then traditional MaxEnt gives $P^{me}_M(X = 1) = 0.1$ which seems a dangerous guess - if it is wrong, it will lead to very bad predictions. In contrast, minimax $P^{me}_M$ gives $P^{me}_M(X = 1) = 0.5$ (one can show that it coincides with the traditional MaxEnt distribution over the convex hull of $\mathcal{C}$) which – to us – seems more reasonable.

**Related Work** (Haussler, 1997) has given a related (but still essentially different) minimax result involving logarithmic *regret* rather than *loss*. (Halpern & Koller, 1995) note that MaxEnt can be made representation independent for a restricted class of representation shifts by restricting the class of priors $\mathcal{M}$ in a certain way, but they do not use this to distinguish between different uses (guesses of $E_{P^*}[\psi]$ for different $\psi$) of the same $P^{me}_M$.

**Acknowledgments** The author is deeply indebted to Phil Dawid, who helped in many crucial ways.